\documentclass[pmlr,twocolumn,10pt]{jmlr} 

\usepackage{longtable}
\usepackage{amsfonts}
\usepackage{algorithm}
\usepackage{algorithmic}
\usepackage{array}
\usepackage{multirow}
\usepackage{rotating}
\usepackage{subcaption}
\usepackage{makecell}

\usepackage{booktabs}
\usepackage{siunitx}

\usepackage[switch]{lineno}

\jmlrvolume{LEAVE UNSET}
\jmlryear{2024}
\jmlrsubmitted{LEAVE UNSET}
\jmlrpublished{LEAVE UNSET}
\jmlrworkshop{Machine Learning for Health (ML4H) 2024} 

 \title{Fairness-Optimized Synthetic EHR Generation for Arbitrary Downstream Predictive Tasks}

\author{%
\Name{Mirza Farhan Bin Tarek} \Email{mfarhan@udel.edu} \\ 
\addr University of Delaware, Newark, DE, USA
\AND
\Name{Raphael Poulain} \Email{rpoulain@udel.edu} \\ 
\addr University of Delaware, Newark, DE, USA
\AND
\Name{Rahmatollah Beheshti} \Email{rbi@udel.edu}\\
\addr University of Delaware, Newark, DE, USA
}

\begin{document}

\maketitle

\begin{abstract}
Among various aspects of ensuring the responsible design of AI tools for healthcare applications, addressing fairness concerns has been a key focus area. Specifically, given the wide spread of electronic health record (EHR) data and their huge potential to inform a wide range of clinical decision support tasks, improving fairness in this category of health AI tools is of key importance. While such a broad problem (mitigating fairness in EHR-based AI models) has been tackled using various methods, task- and model-agnostic methods are noticeably rare. In this study, we aimed to target this gap by presenting a new pipeline that generates synthetic EHR data, which is not only consistent with (faithful to) the real EHR data but also can reduce the fairness concerns (defined by the end-user) in the downstream tasks, when combined with the real data. We demonstrate the effectiveness of our proposed pipeline across various downstream tasks and two different EHR datasets. Our proposed pipeline can add a widely applicable and complementary tool to the existing toolbox of methods to address fairness in health AI applications, such as those modifying the design of a downstream model.
\end{abstract}

\paragraph*{Data and Code Availability}

We have used the MIMIC-III dataset \citep {johnson2016mimic} and the PIC (Pediatric Intensive Care) dataset \citep{zeng2020pic} in our research. Both are available in Physionet data bank \citep{goldberger2000physiobank}. The codebase for our project is available at  \url{https://github.com/healthylaife/FairSynth} 

\paragraph*{Institutional Review Board (IRB)}

This research work did not require IRB approval.

\section{Introduction}
\label{sec:intro}
Integrating ML and AI tools into healthcare has become increasingly feasible with the widespread adoption of large-scale electronic health record (EHR) systems, which contain rich information related to various medical concepts, including diagnoses, procedures, medications, and lab measurements. Different ML models are helping in various aspects of healthcare, for example, clinical predictive modeling \citep{choi2016retain,choi2017gram, Ramazi2019Multi-ModalProgression,gupta2022obesity}, subtyping patients \citep{mottalib2023subtyping} and treatment suggestions \citep{wang2018supervised,shang2019pre, shang2019gamenet}. These models present both opportunities and challenges in healthcare systems.  For example, while ML models can help with more effective diagnosis and treatment, they also risk exacerbating health inequities among historically marginalized groups,  if they are not carefully designed\citep{dressel2018accuracy,ganev2022robin} or the data used for training has inherent bias \citep{abroshan2022counterfactual}. 

To achieve fairness in clinical ML applications, the literature has presented a diverse array of fairness notions and corresponding methodologies that the user can choose to make the model output fair according to the desired characterization of fairness \citep{mehrabi2021survey}. Synthetic data generation techniques present a novel opportunity to address fairness considerations at the data level, especially in the case of underrepresented populations. This approach enables the creation of datasets that inherently embody fairness properties, potentially ensuring that subsequent models trained on these data will exhibit fair behavior in arbitrary downstream tasks. The primary objective of fair synthetic data generation is to produce a dataset that closely approximates the original data distribution while simultaneously mitigating or eliminating discriminatory biases present in the source data. It is important to note that this process of fairness-aware synthetic data generation inevitably alters the underlying data distribution. Consequently, models trained on such synthetically generated fair datasets may experience a degree of performance degradation compared to those trained on the original, unmodified data. One alternative option to using only synthetic data or naively adding synthetic data to the existing data (e.g., for oversampling) for fairness mitigation is generating customized synthetic data that especially targets addressing fairness when augmented to (merged with) the existing data. 


This study aims to propose an end-to-end pipeline for the above purpose, which is generating on-demand synthetic data to mitigate fairness concerns in arbitrary downstream tasks using EHR data. Our focus is distinct from prior work on synthetic EHR generation\citep{abroshan2022counterfactual, Theodorou2023}, as we target not the synthetic data alone, but its combination with existing real data for usage in the user-defined down-stream ML tasks. Additionally, we shift away from the common notion of fair data generation\citep{xu2018fairgan, XuFairGanPlus2019}, as fairness (a sociotechnical phenomenon) is greatly context-dependent, and hence fair synthetic data by itself may have a limited value.
In particular, our key contributions are:
\begin{enumerate}
    \item  We propose an end-to-end pipeline for mitigation of fairness in downstream health ML tasks, with modules for processing EHR data, generating fairness-optimized (FO) synthetic EHR, and a prediction module. The fairness notion,  the generator, and the prediction module can be user-defined.
    \item 
    Through a series of extensive experiments, we show that our proposed pipeline can effectively help with mitigating fairness in various downstream tasks while maintaining a reasonable trade-off between improving fairness and the performance of the downstream tasks. 
\end{enumerate}

\section{Related Work}

Fairness in ML applications can be viewed in various ways. One common way relates to categorizing fairness into two primary types of: group fairness \citep{chouldechova2018frontiers}, and (ii) individual fairness \citep{dwork2012fairness}. Our focus in this research is primarily on group fairness, which involves assessing the discrepancy in performance across the groups (including the protected and privileged groups). This emphasis aims to highlight systemic biases against specific groups. As defined in the literature, group fairness exhibits diverse interpretations, often selected based on stakeholders' perspectives and, at times, presenting conflicting behaviors in specific contexts \citep{narayanan2018translation, poulain2023fairness}.

Relatedly, another common (despite its caveats) way to categorize the methods to mitigate bias in the ML community includes (i) pre-processing  (data-related) techniques, which modify input data to eliminate information correlated with sensitive attributes\citep{bellamy2018a}; (ii) in-processing (model parameter-related) methods, which impose fairness constraints during the model learning process\citep{d2017conscientious,berk2017convex}; and (iii) post-processing (inference-based) approaches, which adjust model predictions after training \citep{ras2022explainable}. Our research will process the data and optimize a generator model for fairness.

On the EHR side, many approaches exist for generating synthetic EHR data. A common approach is using Generative Adversarial Networks (GANs), which have been used in generating EHR data with moderate success \citep{goodfellow2014generative, choi2017generating, zhang2021synteg, torfi2020corgan, cui2020conan, baowaly2019synthesizing, sun2021generating}. Generating sequential EHR data is challenging with GANs as they often produce individual outputs without temporal connections (at least initially). To address this limitation, various strategies have been employed, such as aggregating data into one-time steps \citep{zhang2020ensuring, yan2020generating,rashidian2020smooth}, creating data representations \citep{cui2020conan}, or combining both approaches \citep{choi2017generating, baowaly2019synthesizing}. GANs also face difficulties handling high-dimensional and sparse real-world EHR data, restricting existing synthetic EHR GAN approaches to relatively low-dimensional output by aggregating visits and medical codes or removing rare codes. For example, methods generating longitudinal data, such as \texttt{LongGAN} \citep{sun2021generating} and \texttt{EHR-M-GAN} \citep{li2023generating}, focus solely on dense lab time series with dimensions under a hundred. 

Another synthetic data generation approach is diffusion models, which have emerged as an approach for realistic synthetic data generation. Specifically, denoising diffusion probabilistic models (DDPMs) are a class of latent variable models that can learn the underlying distribution of data by transforming samples into standard Gaussian noise and training the model to denoise these corrupted samples back to their original forms \cite{ceritli2023synthesizing}. 
Recently, \citet{he2023meddiff} proposed \texttt{medDiff}, a DDPM specifically designed for EHRs, showcasing the benefits of diffusion models over existing methods. However, \texttt{medDiff} works best to generate continuous values as it employs a Gaussian diffusion process, assuming the data follows a Gaussian distribution.



Finally, with the rise of large language models (LLMs), researchers have used them for synthetic EHR data generation. 
\texttt{HALO} or Hierarchical Autoregressive Language Model for synthetic EHR data generator is one example, which can generate a probability density function over different medical codes, clinical visits, and patient records. This allows the generation of realistic EHR data \citep{Theodorou2023}. In addition to autoregressive methods, other decoder-based LLMs, like GPT or Generative Pre-trained Transformer models have been used to generate EHRs. For instance, \citet{pang2024cehrgpt} proposed the \texttt{CEHR-GPT} framework that treats patient generation as a language modeling problem similar to \texttt{HALO}, but they have used GPT for the task. 


\section{Problem Statement}
In our work, we aim to develop a pipeline for achieving less bias in the downstream prediction tasks, and to that goal, we have developed a method for generating Fairness-Optimized (FO)  EHR data, by implementing a fairness objective $F$ in the synthetic data generator and later joining that data and the real data together to create an augmented dataset which can help in mitigating bias in downstream tasks. Now, we establish a formal representation of  our pipeline below:

Given a real EHR dataset $R$, a fairness objective function $F$, and a generation process function $G$, the goal of the pipeline is to reduce bias in downstream tasks by generating an FO synthetic EHR dataset $R'$ to augment with $R$. Accordingly, we represent our pipeline as:
\begin{equation}
\label{eq1}
    R \cup G(R, F) = R \cup R'.
\end{equation}

We represent the fairness objective $F$ with a loss function $\mathcal{L}_F$, weighted by the fairness weight (budget) $\lambda$, which is added to the synthetic data generator's loss $L(y,\hat{y})$ to calculate the total loss $L_{model}$: 

\begin{equation}
\label{eq2}
    \mathcal{L}_{model} =  \mathcal{L}(y,\hat{y}) + \lambda \mathcal{L}_F.
\end{equation}

\section{Pipeline Description}
\begin{figure*}[t]
    \centering
    \includegraphics[width=\textwidth]{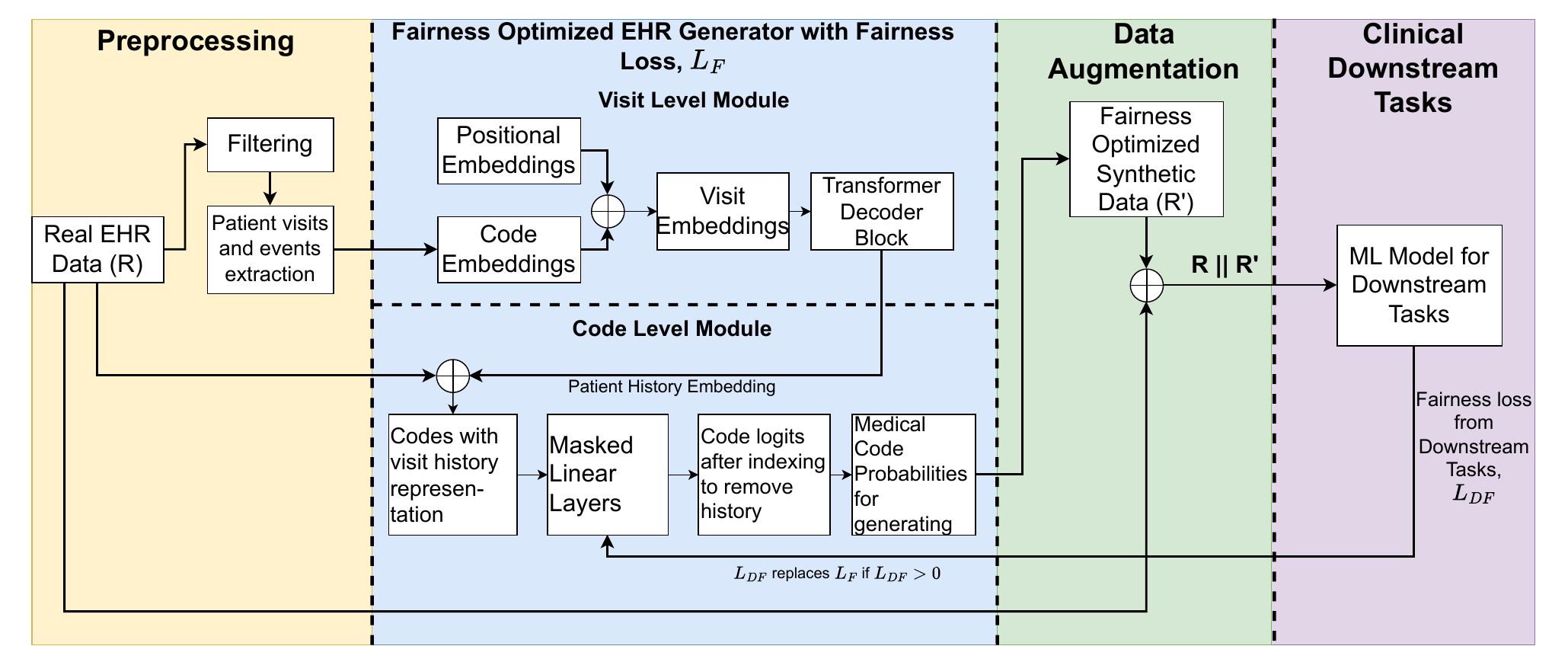}
    \caption{Our proposed pipeline comprises three main components: data pre-processing, generation, and augmentation. The pre-processing stage prepares EHR data for the generator model. We adopt and extend the \texttt{HALO} model with fairness optimization for generating fair synthetic EHR data. The generator features a visit-level module processing visit embeddings and a code-level module ensuring intra-visit cohesion. The generator's loss function incorporates a fairness objective, with a fairness metric from the downstream task fed back.}
    \label{fig: diagram} 
\end{figure*}

Our proposed pipeline consists of three parts: i) data pre-processing, ii) FO synthetic data generation, and iii) data augmentation (Figure \ref{fig: diagram}). In the following, we describe these three parts.

\subsection{Data Pre-processing}
Pre-processing includes filtering the data based on specific features and imputing the missing values. Here, we process the input EHR data in a format suitable for the generator $G$. The data $R$ comprises $N$ patient records $r$, represented as a series of visits $V$, over $T$ time steps. We represent the $i$-th patient record as $r_i = \{V_1^i, V_2^i, V_3^i,\dots, V_T^i\}$. Each visit $V_t$ also contains several medical codes $m_1^t, m_2^t,\dots, m_{k}^t \in C$, labs $l_1^t, l_2^t,\dots,l_{n}^t \in L$ where $k = |V_C^{(t)}|$ and $n = |V_L^{(t)}|$, $C$ is the set of medical codes (diagnosis, procedures and medications) and $L$ is the set of lab tests. 

The pipeline allows users to filter the data based on specific criteria and output preprocessed data. The patient records also contain the labels $D$ for the disease phenotypes and static information $S$ for each patient, possibly including a certain sensitive attribute $s_a \in [[1, M]]$, where $s_a$ is a categorical sensitive attribute with $M$ possible values (e.g. gender, age, race, and ethnicity).

\subsection{Fairness-Optimized Synthetic Data Generation}

We adopt the \texttt{HALO} EHR generator (represented as $G$) and extend that in our pipeline. \texttt{HALO} uses visit-level and code-level modules for generation. The visit-level module processes a patient's medical history using transformer decoder blocks, generating visit history representations that summarize the patient's trajectory. The code-level module generates each variable within a visit based on the patient's history and previously generated variables, ensuring intra-visit cohesion. 

Our pipeline takes the concatenated visit history embeddings and the preprocessed data $R$ as input to create a representation of the codes with visit history, fed through masked linear layers to model the distribution of each patient code, preserving the autoregressive property. The final output represents the code probabilities used for generating synthetic data.


To define the FO objective function for the generation process, let us represent the predicted probabilities for each medical code as $P$, where  $P \in \mathbb{R}^{N \times T \times C}$. Let us also denote $M$ as EHR masks, where $M \in \{y_1,y_2,\dots\}^{N \times T \times C}$.

For model loss $\mathcal{L}(y,\hat{y})$, we implement the binary cross-entropy (BCE) loss which measures the dissimilarity between the predicted probabilities and the ground truth labels, as follows,

\begin{equation}
    \mathcal{L}_{BCE} = -\frac{1}{N} \sum_{i=1}^N \sum_{j=1}^M y_{ij} \log (p_{ij}),
\end{equation}

\noindent where $N$ is the number of patients and $M$ is the number of classes. Specifically, the binary cross entropy works on the medical codes recorded on a patient's visits and whether a certain medical code exists or not in a certain visit.
We can now rewrite Eq \ref{eq2} as:

\begin{equation}
\label{eq4}
    \mathcal{L}_{model} =   \mathcal{L}_{BCE} + \lambda\mathcal{L}_F.
\end{equation}

The generalized FO objective function, $\mathcal{L}_{F}$, quantifies the disparity in the predicted probabilities between different subgroups of the sensitive attribute. Based on the fairness task $F$, a user wants to tackle, it inputs the probabilities of the appearance of the medical codes in a visit, $P$ and the sensitive attribute labels $S$.

\begin{algorithm}
\small
\caption{Fairness Optimization (FO) Objective}
\label{alg:fairness_loss}
    \begin{algorithmic}[1]
    \REQUIRE $P \in \mathbb{R}^{N \times C}$ - Predicted probabilities for each outcome, where $N$ is the number of patients and $C$ is the number of possible outcomes
    \REQUIRE $S \in {1, 2, ..., K}^N$ - Sensitive attribute labels for each patient, where $K$ is the number of subgroups
    \REQUIRE $F$ is a fairness objective
    \ENSURE $\mathcal{L}_{F}$, Fairness loss
    \STATE $\mathcal{S} \leftarrow \text{unique}(S)$ \COMMENT{Get the unique sensitive attribute subgroups}
    \STATE Initialize fairness metric-specific variables
    \FOR{$i \leftarrow 1$ \TO $N$}
    \STATE $s \leftarrow S_i$ \COMMENT{Get the sensitive attribute subgroup of instance $i$}
    \STATE $j \leftarrow \text{index}(s, \mathcal{S})$ \COMMENT{Find the index of the subgroup in $\mathcal{S}$}
    \STATE Update fairness metric-specific variables based on $P{i}$ and $j$
    \ENDFOR
    \STATE Calculate fairness metric-specific values
    \STATE $\mathcal{L}_{F} \leftarrow$ Compute fairness loss based on the fairness metric
    \RETURN $\mathcal{L}_{F}$
    \end{algorithmic}
\end{algorithm}




Algorithm \ref{alg:fairness_loss} computes $\mathcal{L}_{F}$ using the predicted probabilities ($P$) and the sensitive attribute labels. The user specifies the fairness objective to address a particular bias. 

The procedure first identifies the unique subgroups $S_K$ within the sensitive attribute and initializes variables to calculate the fairness loss such as subgroup statistics). It then iterates over each instance, updating these variables based on the predicted probabilities $P_i$ for the instance and its corresponding subgroup. After processing all instances, the algorithm calculates  $\mathcal{L}_{F}$ using the accumulated values, which act as regularization terms in the model's overall loss during training.

One can also input the desired fairness metric of a downstream prediction task (for example, disparate impact in the mortality prediction task) into the model loss calculation as a kind of feedback. We call this value $\mathcal{L}_{DF}$. If $\mathcal{L}_{DF} > 0$, then the model loss would be the sum of $\mathcal{L}_{BCE}$ and $\mathcal{L}_{DF}$ as we have calculated the value of $\mathcal{L}_{DF}$ from the downstream task. This ensures that the model is optimized directly based on the fairness performance in the specific downstream task of interest. Combining both $\mathcal{L}_{F}$ and $\mathcal{L}_{DF}$ in the calculation of model loss $\mathcal{L}_{model}$, we represent the final equation for the loss calculation as:

\begin{equation}
\label{eq5}
\centering
\mathcal{L}_{model} = 
    \begin{cases}
    \mathcal{L}_{BCE} + \lambda \mathcal{L}_{DF}, \text{when } {L}_{DF} > 0\\
    \mathcal{L}_{BCE} + \lambda \mathcal{L}_{F}, \text{otherwise}.
    \end{cases}
\end{equation}

The fairness weight $\lambda$ controls the trade-off between the model's predictive performance (measured by the BCE loss) and its fairness to the sensitive attribute (measured by the generalized fairness loss).

Model training through minimizing the above total loss $\mathcal{L}$ encourages the model to learn to make accurate predictions (by minimizing the BCE loss) while simultaneously reducing the disparity between sensitive attribute subgroups (by minimizing the generalized fairness loss). The goal is to balance predictive performance and fairness, as determined by the fairness weight $\lambda$.

\subsection{Augmentation of Real Data}
After we generate the FO synthetic data $R'$ and ensure that the real EHR data $R$ and the FO synthetic data are in the same format, we merge both of the datasets to finally obtain $R+R'$. Note that while $R$ is fixed throughout the process, $R'$ is iteratively updated to find the optimized $R+R'$ data pool, where `optimized' is determined through user-specified fairness objective in the context of downstream task (for instance, minimizing the worst TPR). The final merged pool of real and synthetic data can be then used in the downstream prediction task. 

\section{Experiments}

\paragraph{Data Description} We perform our  experiments over the MIMIC-III\citep{johnson2016mimic} and Pediatric Intensive Care (PIC)\citep{zeng2020pic} datasets. We extract two cohorts from these two datasets. We present additional details about our data sources and extraction in Appendix \ref{apd:data}.

We extracted time-series events, episode-level information (age, ethnicity), and patient mortality from both the MIMIC-III and PIC dataset following the procedure described by \citet{harutyunyan2019multitask}. In both cases, the respective cohort consists of ICU patients and includes both static data (e.g., gender, ethnicity, diagnosis and procedure codes, and expiry flag) and continuous data (e.g., lab results, age). The continuous data from both datasets were converted to discrete values using granular discretization, and static data were converted into labels. The downstream task for both cohorts is to predict mortality in a patient's next visit based on the current visit.

\paragraph{Fairness Characterization}

We conduct our experiments using ethnicity as the sensitive attribute to evaluate our pipeline's effectiveness. 
Our pipeline allows the end-users to select arbitrary fairness metrics for optimization. In our experiments, we focus on two primary ways of measuring fairness related to group (parity-based) and worst-case (minimax) fairness. From the first type, we focus on disparate impact (DI) as one fairness metric (see Appendix-\ref{apd:second} for details). DI compares the selection rates between unprivileged and privileged groups, where an ideal value of 1 indicates fair treatment without bias. 

Additionally, we assess the minimax fairness by calculating the worst-performing true positive rate (WTPR). Best worst-case metrics may be more appropriate than group fairness for medical applications such as diagnosis, as they better satisfy the principles of beneficence and non-maleficence  (i.e., do the best and do no harm).  Specifically, WTPR reflects the model's ability to correctly identify positive cases in the most disadvantaged group \citep{diana2021minimax}. Higher WTPR values indicate better subgroup performance \citep{poulain2023fairness}.

\paragraph{Downstream Prediction Model} Both of these metrics were evaluated on two separate transformer-based prediction models trained on multiple subsets of the real and synthetic datasets. The model's embedding layer holds the embeddings for input features: visit codes, disease labels, and demographics. The final output is a single value denoting whether a patient will be alive in the next hospital visit. We present additional technical details about our downstream model in Appendix \ref{apd:model}.

\paragraph{Baselines} We compare our pipeline (\textit{Real+FairSynth}) to three other comparable approaches to work with the downstream prediction model.  Using these approaches we created eight curated datasets, six of which are related to the baseline datasets. Among the eight datasets, four of them include patients from the MIMIC-III dataset and the rest include patients from the PIC dataset. The four approaches are as below: 

\begin{itemize}
\item \textit{Real-Only}, where only the data from the real dataset is used to train the downstream models.
\item \textit{Real+Synth}, where the real data is augmented with synthetic data generated through the synthetic generator without implementing the FO part.
\item \textit{Real+Oversample}, where the real data is over-sampled using the popular double prioritized bias correction method \citet{Pias2023}. The double prioritized (DP) bias correction method addresses data imbalance in a specific subgroup, such as a particular ethnicity, by incrementally replicating data points of the minority iteratively. In our experiments, we replicated patient data belonging to each minority ethnicity so that the distribution is less skewed.
\item \textit{Real+FairSynth (Our Method)}, where the real data is augmented with the synthetic data that is generated using our proposed pipeline.
\end{itemize}

In our experiments, we also evaluate the impact of the size of the real dataset and the synthetic dataset on the fairness and accuracy of the prediction model.

\section{Results}
\tableref{tab:metrics_all} shows the results obtained through various settings of our experiments across two datasets (MIMIC and PIC), two fairness metrics (DI and WPTR), four approaches, and various sizes of real and synthetic data. 

\paragraph{Variable size of real data, plus fixed synthetic data samples}
Looking at our DI metric (MIMIC data, top table), our proposed method achieved the best results for 2,500 and 5,000 real samples, whereas oversampling the real data achieved better results for fewer samples. Turning to the WTPR metric, where a higher value is better, our proposed method proves to better at 2,500 and 5,000 real samples. It achieves a TPR of .75 and .78, respectively, considerably higher than the other baselines. Meaning, our method achieves high true positive even in the worst cases.

In the case of the PIC dataset (bottom table), for 1,000 and 2,500 data samples, the DI measure was near 1 compared to the other baselines. Shifting focus to the WTPR, Real+FairSynth demonstrates great performance for 1,000 real samples. For example, with 1,000 real samples, Real+FairSynth achieves a WTPR of .29, considerably higher than Real-Only (.12), Real+Oversample (.08), and Real+Synth (.13). However, the Real+Oversample achieved better results for other cases than the proposed method. We suspect that the size of the original PIC dataset has impacted the EHR generative model (on which FairSynth is based) in a negative way, which has affected the proposed method's performance in larger datasets.

\begin{table*}[htbp]
\floatconts
{tab:metrics_all}
{\caption{Combined Results for Real and Synthetic Samples from MIMIC-III (top table) and PIC datasets (bottom table). Mean (SD) values are shown.  R: Real, Over: Over Sample.}}
{%
    \resizebox{\textwidth}{!}{
    \begin{tabular}{@{}l l | l l l l | l l l l | l l l l@{}}
    \toprule
    \multicolumn{2}{c}{Sample Size} & \multicolumn{4}{c}{Disparate Impact (closer to 1 is better)} & \multicolumn{4}{c}{WTPR$\uparrow$} & \multicolumn{4}{c}{F1-Score$\uparrow$} \\
    \cmidrule(r){1-2} \cmidrule(lr){3-6} \cmidrule(lr){7-10} \cmidrule(l){11-14}
    {Real} & {Synth} & {R} & {R+Over} & {R+Synth} & {R+FairSynth} & {R} & {R+Over} & {R+Synth} & {R+FairSynth} & {R} & {R+Over} & {R+Synth} & {R+FairSynth} \\
    \cmidrule(r){1-2} \cmidrule(lr){3-6} \cmidrule(lr){7-10} \cmidrule(l){11-14}
    1,000 & 2,500 & .98(.02) & \textbf{.99(.03)} & .97(.04) & .39(.66) & \textbf{.48(.07)} & .26(.01) & .17(.03) & .44(.23) & \textbf{.53(.06)} & .27(.09) & .25(.03) & .45(.02) \\
    2,500 & 2,500 & .98(.05) & \textbf{.98(.04)} & .97(.04) & 1.20(.20) & .51(.06) & .24(.07) & .17(.01) & \textbf{.75(.05)} & \textbf{.51(.09)} & .28(.07) & .27(.06) & .49(.01) \\
    5,000 & 2,500 & \textbf{.99(.02)} & .99(.02) & .98(.03) & 1.10(.09) & .67(.06) & .14(.02) & .14(.02) & \textbf{.78(.07)} & \textbf{.55(.03)} & .28(.06) & .27(.05) & .46(.02) \\
    \midrule
    2,500 & 500 & {-} & .98(.03) & .98(.04) & \textbf{.98(.22)} & {-} & .16(.04) & .16(.02) & \textbf{.24(.14)} & {-} & .27(.03) & .25(.05) & \textbf{.48(.04)} \\
    2,500 & 1,000 & {-} & \textbf{.99(.03)} & .98(.02) & 1.28(.42) & {-} & .19(.03) & .15(.01) & \textbf{.72(.02)} & {-} & .29(.03) & .26(.04) & \textbf{.50(.03)} \\
    2,500 & 2,000 & {-} & 1.02(.04) & \textbf{.99(.06)} & 1.03(.44) & {-} & .25(.01) & .15(0) & \textbf{.83(.06)} & {-} & .33(.02) & .26(.04) & \textbf{.49(.01)} \\
    \bottomrule
    \end{tabular}}}

\vspace{0.5cm}

{%
\resizebox{\textwidth}{!}{
\begin{tabular}{@{}l l | l l l l | l l l l | l l l l@{}}
\toprule
\multicolumn{2}{c}{Sample Size} & \multicolumn{4}{c}{Disparate Impact$\uparrow$} & \multicolumn{4}{c}{WTPR$\uparrow$} & \multicolumn{4}{c}{F1-Score$\uparrow$} \\
\cmidrule(r){1-2} \cmidrule(lr){3-6} \cmidrule(lr){7-10} \cmidrule(l){11-14}
{Real} & {Synth} & {R} & {R+Over} & {R+Synth} & {R+FairSynth} & {R} & {R+Over} & {R+Synth} & {R+FairSynth} & {R} & {R+Over} & {R+Synth} & {R+FairSynth} \\
\cmidrule(r){1-2} \cmidrule(lr){3-6} \cmidrule(lr){7-10} \cmidrule(l){11-14}
1,000 & 2,500 & .87(.51) & 1.41(.81) & .42(0) & \textbf{.66(.23)} & .12(.03) & .08(.03) & .13(0) & \textbf{.29(.09)} & .38(.25) & \textbf{.62(.02)} & .14(.12) &  .03(.04) \\
2,500 & 2,500 & .98(.05) & .39(.49) & 1.38(.32) & \textbf{.85(.15)} & .06(.05) & .1(.04) & \textbf{.27(.16)} & .06(.05) & .36(.25) & \textbf{.6(.05)} & .12(.06) & .10(.07) \\
5,000 & 2,500 &.76(.77) & \textbf{.95(.18)} & 1.11(.76) & .80(.14) & .05(.02) & .15(.13) & \textbf{.19(.17)} & .15(.20) & .35(.25) & \textbf{.57(.01)} & .09(.05) & .07(.06) \\
\midrule
2,500 & 500 & {-} & 1.34(.44) & 1.3(.52) & \textbf{.6(0)} & {-} & .06(.03) & .1(.6) & \textbf{.03(.03)} & {-} & \textbf{.56(.07)} & .51(.17) & .06(.06) \\
2,500 & 1,000 & {-} & \textbf{.99(.03)} & .98(.02) & 1.28(.42) & {-} & .13(.11) & .04(.01) & \textbf{.02(.04)} & {-} & \textbf{.57(.06)} & .45(.02) & .04(.09) \\
2,500 & 2,000 & {-} & 1.02(.04) & \textbf{.99(.06)} & 1.03(.44) & {-} & .05(.02) & .02(0) & \textbf{.13(.16)} & {-} & \textbf{.58(.05)} & .41(.02) & .06(.08) \\
\bottomrule
\end{tabular}}}
\end{table*}

\paragraph{Variable synthetic data samples and fixed real data samples}
Experiments were also conducted where 2,500 real data samples were augmented using various synthetic samples. Table \ref{tab:metrics_all} (top) shows that increasing the number of synthetic samples yields improvements for all methods. However, the proposed Real+FairSynth consistently achieves DI values nearest to the ideal value of 1. At 500 synthetic samples, Real+FairSynth has a DI of .98. This advantage is maintained at 2,000 synthetic samples, with Real+FairSynth reaching a DI of 1.03 at the 2,000 sample mark, compared to 1.02 and .99 for the other methods.

Turning attention to the WTPR, the results again favor the Real+FairSynth approach as the number of synthetic samples increases. At 500 synthetic samples, Real+FairSynth achieves a WTPR of .24, compared to the results from the other baseline datasets. Finally, increasing the number of synthetic data samples increases the total sample size, and thus, we get better WTPR scores, which is reflected in the scores for 1,000 and 2,000 synthetic samples (.72 and .83). For the PIC dataset, we see similar results.

To summarize, our analysis (\tableref{tab:metrics_all} shows that our method (Real+FairnSynth) achieves best performance in 13 out of 24 different experiments while being the second best in 4 experiments (24 is the number of four-cell rows under DI and WTPR columns). Our method works especially best in the case of WTPR (top method in 9 out of 12). Our results also show that, even with a few real samples, adding synthetic samples generated by our proposed method can improve the fairness metrics (as evidenced by 13 experiments).

These findings suggest that augmenting a fixed real dataset with the increasing number of synthetic samples can enhance fairness metrics and that the proposed Real+FairSynth method is particularly effective.

\subsection{Prediction Performance}
We also studied the prediction performance of our method (Real+FairSynth) using the F1-score for all four approaches. From \tableref{tab:metrics_all} (right part), one can observe the F1-scores for the real and augmented datasets. The table shows results for both the MIMIC-III and PIC datasets across different proportions of real and synthetic samples. While our fairness mitigation method (expectedly) does not achieve the highest predictive performance, it is still noticeable that it maintains proximity to the baselines that do not take fairness considerations into account.

\subsection{Trade-off Between Fairness and Prediction Accuracy}
\label{subsec:trade-off}

To further investigate the trade-offs between prediction performance and fairness metrics, we also run separate experiments by varying the value of the fairness weight (fairness budget) parameter, $\lambda$. This analysis aimed at understanding how different values of $\lambda$ impact both the F1-scores and fairness metrics such as DI and WTPR.

\subsubsection{Impact on Prediction Performance}

\tableref{tab:all_lambdas} illustrates the F1-scores for different values of $\lambda$ on the MIMIC-III dataset (left values on the mean(sd) pairs reported in each cell), across three different sample sizes (1,000, 2,500, and 5,000). As $\lambda$ increases from .5 to 1.5, there is a noticeable decline in the F1 scores. For example, for 5,000 samples, the F1-score reduces from .46 at $\lambda$ = .5 to .24 at $\lambda$ = 1.5. This trend clearly shows that higher values of $\lambda$ negatively impact the prediction performance, as the declining F1 scores indicate. However, for $\lambda$ = 1.2, we get comparatively better F1-score than $\lambda=.5$ or $\lambda=1.5$.

 However, for the PIC dataset in \tableref{tab:all_lambdas} (right values in the cells), one can observe that our proposed method achieved various prediction performance for different numbers of real samples for $\lambda$ values ranging from .5 to 1.5. This is possibly because the original data was heavily skewed towards a certain subpopulation (as shown in Appendix \ref{apd:expo}), which could have affected the generative model's performance negatively, resulting in diverse quality generated data. The current set of results is obtained using a fixed $\lambda$ value for all experiments. In the next part, we show that modifying $\lambda$ would allow maintaining the top performance across different scenarios. 
 

\subsubsection{Impact on Fairness Metrics}
\tableref{tab:all_lambdas} presents the DI and WTPR values for the same variations in $\lambda$ and sample sizes for the MIMIC-III dataset. Here, a higher value of $\lambda$ generally enhances the fairness metrics. However, increasing the $\lambda$ too much, from 1.2 to 1.5, affects the DI impact positively but WTPR negatively, indicating that changing the lambda value causes variation in the fairness metrics. While trying to balance the F1-score and the fairness metrics, it seems that   $\lambda$=1.2 achieves the best overall results.

\begin{table}[htbp]
\caption{Performance on different $\lambda$ values on MIMIC-III and PIC datasets. In each cell, the values are shown as mean(sd) for Mimic-III then (comma) PIC. For DI closer to 1 is better.}
\label{tab:all_lambdas}
\resizebox{1\linewidth}{!}{
    \begin{tabular}{@{}l*{5}{c}@{}}
    \toprule
    & \textbf{\# } & \textbf{$\lambda=0.5$} & \textbf{$\lambda=1.0$} & \textbf{$\lambda=1.2$} & \textbf{$\lambda=1.5$} \\
    \midrule
    & 1,000  & 1.3(1.21),.04(.08) & 1(.05),1(.09) & .89(.66),.03(.04) & 1(.06),.07(.10)\\
    DI & 2,500  & .99(.22),.05(0) & 1.28(.42),.13(.09) & 1.2(.28),.1(.07) & .99(.02),.02(.05) \\
    & 5,000  & 1.1(.09),.06(.05) & .99(.16),.13(.03) & 1.1(.09),.07(.06) & .99(.04),.04(.06) \\
    \midrule
    & 1,000  & .64(.24),1.28(1.21) & .72(.03),1(.05) & .44(.23),.89(.66) & .15(.08),1(.06) \\
    WTPR$\uparrow$ & 2,500  & .74(.09),.99(.22) & .72(.02),1.28(.42) & .75(.05),1.20(.28) & .15(.01),.99(.02) \\
    & 5,000  & .78(.07),1.10(.09) & .73(.08),.99(.16) & .78(.07),1.10(.09) & .15(.02),.99(.04) \\
    \midrule
    & 1,000  & .55(.03),.64(.24) & .10(.09),.72(.03) & .45(.02),.44(.23) & .22(.08),.15(.08) \\
    F1-score$\uparrow$ & 2,500  & .52(.02),.74(.09) & .13(.09),.72(.02) & .49(.01),.75(.05) & .24(.01),.15(.01) \\
    & 5,000  & .46(.02),.78(.07) & .13(.03),.73(.08) & .45(.03),.78(.07) & .24(.02),.15(.02) \\
    \bottomrule
    \end{tabular}}
\end{table}


In summary, we observed a trade-off between accuracy (F1-scores) and fairness (DI and WTPR) in model optimization. Lower $\lambda$ values prioritize prediction performance, leading to higher F1 scores but potentially less fairness, while higher $\lambda$ values improve fairness at the cost of accuracy. We chose $\lambda=1.2$ for the earlier experiments as it strikes a balance between performance and fairness. This trade-off suggests that the choice of $\lambda$ should align with the application's goals, balancing the need for both effective and fair models. Our experiments confirm that in different configurations, adjusting our fairness parameter would help the proposed pipeline achieve the best results in specific scenarios.

\subsection{Limitations}
Our proposed pipeline has several limitations. Firstly, while our pipeline in the current version is limited to tabular data, it still establishes a valuable foundation for generating FO synthetic data and has shown good results. Including other modalities like medical images or medical notes in future versions might improve the pipeline's output. 

Furthermore, we did not train the generator and downstream prediction models on very large-scale EHRs. We expect that our model would maintain its performance on larger datasets and running the models for a longer time would improve the output quality even further. Finally, our current pipeline version supports augmentation as the sole debiasing method. Other types of fairness mitigation methods (like inference-time methods) can be merged with our pipeline.

\section{Conclusion}
In this study, we presented a new pipeline for mitigating bias in clinical downstream tasks with the help of fairness-optimized (FO) augmented EHR data. Specifically, we used the FO objective function to ensure that combining the real data with the generated synthetic data can improve the fairness performance of downstream predictive models running on the data. Through a series of comprehensive experiments on MIMIC-III and pediatric ICU datasets, we showed that adding FO objectives in the generator and augmenting real data with FO synthetic data can reduce the biases observed for different sensitive attributes in downstream mortality prediction tasks.  Even though our method solely focuses on data-related methods where we try to improve fairness by augmenting, combining our proposed method with model-related techniques (where the downstream model is modified) and inference-related techniques (where post-processing is performed on the output) is possible. 


\bibliography{ml4h}

\appendix
\section{Transformer Decoder Block}
\label{apd:first}
The visit-level transformer module described in the main work utilizes a stack of Transformer Decoder blocks from the original Transformer paper \citep{vaswani2017attention}. Additional details on those blocks are given below. Each block is defined mathematically by

\begin{equation}
\begin{split}
\mathbf{H}_1^{(m)} &= \mathbf{H}^{(m-1)} +
\text{MMSA}(\mathbf{H}^{(m-1)}) \\
\mathbf{H}_2^{(m)} &= \text{Layer Normalization}(\mathbf{H}_1^{(m)}) \\
\mathbf{H}_3^{(m)} &= \mathbf{H}_2^{(m)} + \text{max}(0, \mathbf{H}_2^{(m)} \mathbf{W}^{(m)} + \mathbf{b}^{(m)} + \\
\mathbf{c}^{(m)})\mathbf{V}^{(m)}) \\
\mathbf{H}^{(m)} &= \text{Layer Normalization}(\mathbf{H}_3^{(m)})
\end{split}
\end{equation}

\noindent
where Masked Multi-Head Self-Attention (MMSA) is then defined by

\begin{equation}
\begin{split}
\text{MMSA}(\mathbf{V}) = \text{Concat}(\text{head}_1, \cdots, \text{head}_h)\mathbf{W}^O \\
\text{head}_i = \text{Masked Attention}(\mathbf{V}\mathbf{W}_i^Q, \mathbf{V}\mathbf{W}_i^K, \mathbf{V}\mathbf{W}_i^V) \\
\text{Masked Attention}(\mathbf{Q}, \mathbf{K},\\ \mathbf{V}) = \text{softmax}\left(\frac{\mathbf{Q}\mathbf{K}^\top}{\sqrt{d_k}} + \mathbf{M}\right) \mathbf{V}
\end{split}
\end{equation}

with $M$ in the final line being a triangular matrix of $-\infty$ values, the softmax calculation allows elements in the sequence to attend to themselves and elements before them only.

\section{Description of the Fairness Optimization (FO) Objective Functions Used in Experiments}
\label{apd:second}

\textbf{Disparate impact} examines the likelihood of receiving a positive classification. However, instead of focusing on the difference between unprivileged and privileged groups, the ratio of these probabilities is considered. We can formalize DI as follows: 

Let $SR_p$ be the selection rate for the protected group, and $SR_n$ be the selection rate for the non-protected group. The selection rate for a group is calculated as the number of favorable outcomes (e.g., positive predictions, approvals) divided by the total number of instances in that group.

So, the disparate impact ratio is $DI$ then calculated as:

\begin{equation}
DI = \frac{SR_p}{SR_n}
\end{equation}
The implementation is shown in algorithm \ref{alg:di_loss_gen}.

\begin{algorithm}[h]
\caption{Disparate Impact Loss Function}
\label{alg:di_loss_gen}
    \begin{algorithmic}[1]
    \REQUIRE $P \in \mathbb{R}^{N \times C}$ - Predicted probabilities for each code, where $N$ is the number of patients and $C$ is the number of codes
    \REQUIRE $G \in {...N}^N$ - \COMMENT{Subgroup labels for a sensitive attribute}
    \ENSURE $\mathcal{L}_{DI}$ - Disparate impact loss
    \STATE $\mathcal{G} \leftarrow \text{unique}(G)$ \COMMENT{Get the unique subgroup labels}
    \STATE $S_g \leftarrow {0}^{|\mathcal{G}|}$ \COMMENT{Initialize sensitive attribute-wise positive prediction sums}
    \STATE $N_g \leftarrow {0}^{|\mathcal{G}|}$ \COMMENT{Initialize sensitive attribute-wise patient counts}
    \FOR{$i \leftarrow 1$ \TO $N$}
    \STATE $g \leftarrow G_i$ \COMMENT{Get the sensitive attribute value of patient $i$}
    \STATE $j \leftarrow \text{index}(g, \mathcal{G})$ \COMMENT{Find the index of the sensitive attribute in $\mathcal{G}$}
    \STATE $p_i \leftarrow \sum_{c=1}^C \mathbf{1}(P_{i,c} > .5)$ \COMMENT{Count positive predictions for patient $i$}
    \STATE $S_{g_j} \leftarrow S_{g_j} + p_i$ \COMMENT{Add positive prediction count to the corresponding value of the sensitive attribute}
    \STATE $N_{g_j} \leftarrow N_{g_j} + 1$ \COMMENT{Increment the count for the corresponding sub group}
    \ENDFOR
    \STATE $R_g \leftarrow \frac{S_g}{N_g}, \forall g \in \mathcal{G}$ \COMMENT{Calculate the positive prediction rate for each sub group}
    \STATE $DI \leftarrow \frac{R_{g_0}}{R_{g_1}}$ \COMMENT{Compute the disparate impact ratio}
    \STATE $\mathcal{L}_{DI} \leftarrow |1 - DI|$ \COMMENT{Calculate the disparate impact loss}
    \RETURN $\mathcal{L}_{DI}$
    \end{algorithmic}
\end{algorithm}

The loss function inputs the predicted probabilities for each code ($P$) and the sensitive attribute labels for each patient ($G$). It first identifies the unique sensitive attribute labels present in the dataset ($\mathcal{G}$). Then, it initializes two arrays, $S_g$ and $N_g$, to store the sum of positive predictions and the count of patients for each sensitive attribute group, respectively.

The algorithm then iterates over each patient $i$ and determines their subgroup $g$. It finds the corresponding index $j$ of the sensitive attribute in the unique label $\mathcal{G}$. For each patient, the function counts the number of positive predictions ($p_i$) by summing the indicator function $\mathbf{1}(P_{i,c} > .5)$ over all codes $c$. This indicator function returns 1 if the predicted probability $P_{i,c}$ is greater than .5 (i.e., a positive prediction) and 0 otherwise. The count of positive predictions is added to the corresponding sensitive attribute's sum $S_{g_j}$, and the patient count for that sensitive attribute $N_{g_j}$ is incremented.

After processing all patients, each subgroup's positive prediction rate $R_g$ is calculated by dividing the sum of positive predictions $S_g$ by the patient count $N_g$ for each ethnicity $g$. The disparate impact ratio $DI$ is then computed as the ratio of the positive prediction rates between the subgroups groups.

Finally, the disparate impact loss $\mathcal{L}_{DI}$ is calculated as the absolute difference between 1 and the disparate impact ratio $DI$. This loss function quantifies the deviation from perfect equality (i.e., a disparate impact ratio of 1) between the ethnic groups. A higher loss value indicates a larger disparate impact, while a lower loss value suggests a more balanced and fair model.

\textbf{Worst performing TPR (the lowest) or the Worst TPR}. It is a useful fairness metric because it focuses on the model's performance for the most disadvantaged subgroup. It helps identify if there are significant disparities in the model's performance across different subgroups defined by the protected attribute. Suppose $s_a \in [[1, M]]$, where $s_a$ is a categorical sensitive attribute with $M$ possible values (e.g. gender, age, race, and ethnicity), $Y \in[0,1]$ and $\hat{Y}\in[0,1]$ as the ground truth and the prediction of a binary predictor, respectively.
\begin{equation}
\text{Worst-Case TPR} = \min_{m \in M} Pr(\hat{Y}=1|A=m, Y=1)
\end{equation}

A smaller value (closer to zero) is ideal for the parity-based measures, but for the worst TPR, a greater value(closer to one) is desired.

\section{Data Description}
\label{apd:data}
The MIMIC-III dataset is sourced from the Medical Information Mart for Intensive Care (MIMIC) to analyze inpatient records. This extensive dataset encompasses anonymized EHRs from patients admitted to the Beth Israel Deaconess Medical Center in Boston, MA, USA, from 2001 to 2012. The data contains each patient's hospital stay information, including laboratory measurements, medications administered, and vital signs. 
From the MIMIC-III dataset, we have used the ICU stay dataset. It contains 46,520 patients with 25 disease phenotype labels which were defined by the MIMIC benchmark \citep{harutyunyan2019multitask}. Among the patients, 32,950 are male, and the remaining are females. The dataset has more than 30 ethnicities; however, to simplify processing, we decided to group the different ethnicities into 5 categories based on the ethnicity names containing White, Black, Hispanic, or Asian keywords.  The White ethnicity has the most samples in the dataset, which is 41,325 samples, and the rest are of the other groups.  On the other hand, the Pediatric Intensive Care (PIC) database is a large pediatric-specific, single-center, bilingual database comprising information relating to children admitted to critical care units at the Children’s Hospital, Zhejiang University School of Medicine. The dataset has a similar structure to the MIMIC-III dataset; however, the dataset is smaller, having 13,449 patients. Among the patients described in the PIC dataset, 7,728 of are male and the rest are females. Considering the ethnic distribution, it is very skewed toward the Han ethnic group (13,300 patients) and the rest of the patient population belongs to the Miao, Tujia, Buyei, Yi, and Hui ethnic groups.

\section{Exploratory Data Analysis}
\label{apd:expo}
\subsection{MIMIC-III Dataset}
In our MIMIC-III cohort, we have selected 5,000 random patients. Among them, 2807 are males and the rest are female patients. The patients belong to 5 ethnicities: White, Black, Hispanic, Asian, and Others. Among the different ethnicities, the mortality of the patients is shown in the \tableref{tab:mimic_dist}.

\begin{table}[htbp]
\floatconts
{tab:mimic_dist}
{\caption{The number of alive and expired patients of different ethnicities in the MIMIC-III cohort}}
{%
\begin{tabular}{@{}|l|l|l|@{}}
\toprule
 & \textbf{Alive} & \textbf{Expired} \\ \midrule
\textbf{White} & 3105 & 359 \\ \midrule
\textbf{Black} & 460 & 27 \\ \midrule
\textbf{Hispanic} & 175 & 11 \\ \midrule
\textbf{Asian} & 156 & 9 \\ \midrule
\textbf{Others} & 615 & 83 \\ \bottomrule
\end{tabular}
}
\end{table}

\subsection{PIC Dataset}
In our PIC cohort, we have selected 5,000 random patients. Among them, 2832 are males and the rest are female patients. The patients belong to 5 ethnicities: Han, Yi, Buyei, Miao, Tujia, Hui, and Others. Among the different ethnicities, the mortality of the patients is shown in the \tableref{tab:pic_dist}.

\begin{table}[t]
\floatconts
{tab:pic_dist}
{\caption{The number of alive and expired patients of different ethnicities in the PIC cohort}}
{%
\begin{tabular}{@{}|l|l|l|@{}}
\toprule
 & \textbf{Alive} & \textbf{Expired} \\ \midrule
\textbf{Han} & 2302 & 193 \\ \midrule
\textbf{Yi} & 364 & 35 \\ \midrule
\textbf{Buyei} & 393 & 26 \\ \midrule
\textbf{Miao} & 365 & 27 \\ \midrule
\textbf{Hui} & 402 & 27 \\ \midrule
\textbf{Others} & 396 & 39 \\ \bottomrule
\end{tabular}
}
\end{table}

\section{Downstream Model Design}
\label{apd:model}
The model's core is the transformer component, which contains separate transformer layers for each input feature (visit codes, disease labels, ethnicity). Each transformer layer consists of a transformer block with a multi-headed attention module for capturing dependencies within the input sequence and a position-wise feed-forward module for non-linear transformations. The transformer layers also include layer normalization and dropout regularization. The transformer model is described in appendix \ref{apd:first}.

The linear layers include separate linear transformations for visit codes, disease labels, and ethnicity. These linear layers project the input features into a common 128-dimensional space.

The outputs from the transformer layers are concatenated and passed through a fully connected layer to produce the model's output, which is a single value.
The model is trained using an Adam optimizer with a learning rate of .001, no weight decay, and no maximum gradient norm. The batch size is set to 64, and the model is trained for 1 epoch due to limited computing resources.

The training data were built in two ways: i) augmenting a varying number of real data samples with a fixed number of synthetic samples and ii) augmenting a fixed number of real data samples with a varying number of synthetic samples. The whole dataset was divided into 80-10-10 train-validation-test split. 
\section{Experimental Setup}
\label{apd:third}
We split the data into an 80-10-10 ratio for training, validation, and testing. Due to limited computing resources and GPU power, we ran the synthetic EHR generator using the Adam optimizer with a batch size of 10, a sample batch size of 25, and a learning rate of $10^{-4}$. 

All of our experiments were run using the built-in ML models, specifically a transformer-based model, from the PyHealth library. These models utilize static data, such as demographic information e.g. gender and ethnicity, and continuous time-series data.

The experiments were performed on three environments: Two Windows PCs and in the cloud using Google Colab. The first machine had an Intel Core i9-9900K processor, 32 GB of RAM, and an NVIDIA GeForce RTX 2080Ti GPU with 12GB memory. The second machine featured an Intel Core i9-13900HX processor, 32 GB of RAM, and an NVIDIA GeForce RTX 4070 GPU with 8GB memory. In Google Colab, we utilized the T4, L4, and A100 GPUs. All the codes were run using Python version 3.11.5, CUDA version 12.1, and PyTorch 2.1.2.

\end{document}